\pdfoutput=1

\documentclass[11pt]{article}

\usepackage{EMNLP2023}

\usepackage{times}
\usepackage{latexsym}

\usepackage[T1]{fontenc}

\usepackage[utf8]{inputenc}

\usepackage{microtype}

\usepackage{inconsolata}
\usepackage{graphicx}
\usepackage{amsmath}
\usepackage{booktabs, tabularx}
\usepackage{multirow}
\usepackage{tikz}
\usepackage{pgfplots}
\pgfplotsset{compat=1.18}

\usepackage[ruled,vlined]{algorithm2e}
\definecolor{commentcolor}{RGB}{110,154,155}   
\newcommand{\PyComment}[1]{\ttfamily\textcolor{commentcolor}{\# #1}}
\newcommand{\PyCode}[1]{\ttfamily\textcolor{black}{#1}} 

\newcommand{\flag}[1]{#1}
\newcommand{\mycomment}[1]{}
\newcommand{\modelname}{CASE}

\newcommand{\metricname}{$F_1$ }

\newcommand{\declarelogo}[0]{\includegraphics[height=.02\textwidth]{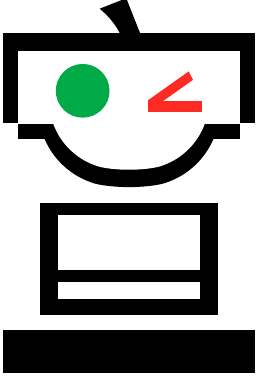}}

\title{Domain-Expanded ASTE: Rethinking\\
Generalization in Aspect Sentiment Triplet Extraction}

\author{
\textbf{
Yew Ken Chia\thanks{~~Yew Ken and Guizhen are students under the Joint PhD Program between Alibaba and their corresponding university. This work is done by Hui and Wei during internship at Alibaba.
}
~\textsuperscript{\rm 1,~${\declarelogo}$}
\quad Hui Chen\textsuperscript{\rm ~${\declarelogo}$}
\quad
Guizhen Chen\textsuperscript{\rm ~1,~2}
\quad 
Wei Han\textsuperscript{\rm ~${\declarelogo}$}
} \\
\textbf{
Sharifah Mahani Aljunied\textsuperscript{\rm ~1} 
\quad
Soujanya Poria\textsuperscript{\rm ~${\declarelogo}$}
\quad 
Lidong Bing\textsuperscript{\rm ~1}
} \\
\textsuperscript{\rm ${\declarelogo}$} Singapore University of Technology and Design ~~\\
\textsuperscript{\rm 1}DAMO Academy, Alibaba Group, Singapore~~  \\
\textsuperscript{\rm 2}Nanyang Technological University, Singapore \\
{\tt sporia@sutd.edu.sg} 
\quad
{\tt guizhen001@ntu.edu.sg} \\
{\tt\{yewken\_chia, hui\_chen, wei\_han\}@mymail.sutd.edu.sg} \\
{\tt\{yewken.chia, guizhen.chen, mahani.aljunied, l.bing\}@alibaba-inc.com}
}

\begin{document}
\maketitle
\begin{abstract}
Aspect Sentiment Triplet Extraction (ASTE) is a challenging task in sentiment analysis, aiming to provide fine-grained insights into human sentiments. However, existing benchmarks are limited to two domains and do not evaluate model performance on unseen domains, raising concerns about the generalization of proposed methods. Furthermore, it remains unclear if large language models (LLMs) can effectively handle complex sentiment tasks like ASTE. In this work, we address the issue of generalization in ASTE from both a benchmarking and modeling perspective. We introduce a domain-expanded benchmark by annotating samples from diverse domains, enabling evaluation of models in both in-domain and out-of-domain settings. Additionally, we propose CASE, a simple and effective decoding strategy that enhances trustworthiness and performance of LLMs in ASTE. Through comprehensive experiments involving multiple tasks, settings, and models, we demonstrate that CASE can serve as a general decoding strategy for complex sentiment tasks. By expanding the scope of evaluation and providing a more reliable decoding strategy, we aim to inspire the research community to reevaluate the generalizability of benchmarks and models for ASTE. Our code, data, and models are available at \url{https://github.com/DAMO-NLP-SG/domain-expanded-aste}.
\end{abstract}

\section{Introduction}

Opinions and sentiments are essential to human communication, beliefs, and behaviors \cite{liu2012sentiment}.
Although sentiment analysis is often performed at the sentence or document level, it is insufficient to capture the fine-grained sentiment information and nuances of human opinions \cite{Poria2020BeneathTT}.
To this end, aspect sentiment triplet extraction (ASTE) \cite{peng2020knowing} is a challenging and well-established task of aspect-based sentiment analysis \cite{pontiki-etal-2014-semeval} which aims to extract richer and more interpretable sentiment information from natural language.
Concretely, ASTE considers how each opinion term in a text may express sentiments towards specific aspect targets.

\begin{figure}[t]
    \centering
    \includegraphics[width=1.0\columnwidth]{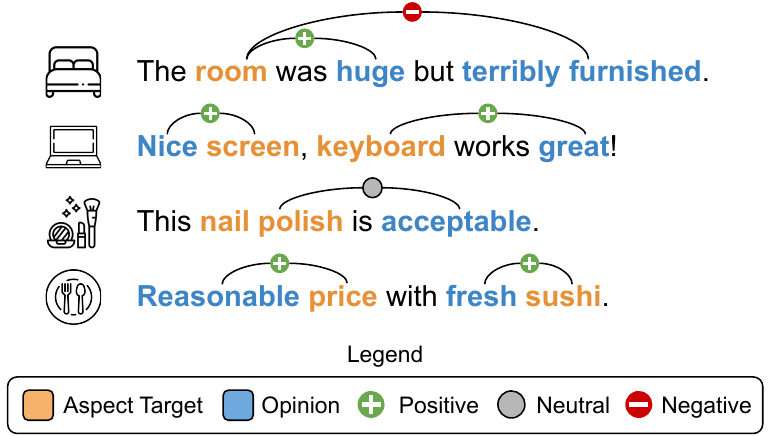}
    \caption{
    ASTE data samples for the Hotel, Laptop, Cosmetics, and Restaurant domains, respectively.
    }
    \label{fig:cover}
\end{figure}

Although ASTE has become a more established task with many existing methods \cite{Zhang2022ASO}, we are concerned that they may not generalize well due to limitations in the existing benchmark datasets.
Notably, the established benchmarks are limited to two domains, which limits the evaluation scope of model capabilities and does not represent the diversity of real-world data.
On the other hand, it is also important to assess how models generalize to unseen domains as domain-specific labeled data is often scarce \cite{wang-pan-2018-recursive}, and models may face 
domain shift during deployment \cite{Wang2021GeneralizingTU}. 
Hence, this motivates us to propose a domain-expanded ASTE benchmark which not only considers the in-domain performance, but also evaluates out-of-domain generalization across a more diverse set of domains.
We support the new benchmark by annotating more than 4,000 data samples for two new domains based on hotel and cosmetics product reviews.
Therefore, we can construct a domain-expanded dataset with four domains as shown in Figure \ref{fig:cover}.

To investigate the domain generalization of existing ASTE methods, we evaluate five existing methods based on pretrained language models (PLMs) for the in-domain and out-of-domain settings.
On the other hand, while large language models (LLMs) have recently enabled breakthroughs in many NLP tasks, it is unclear if they can surpass specialized pretrained language models (PLMs) on sentiment tasks such as ASTE \cite{zhang2023reality}.
Despite the impressive language understanding and general-purpose capabilities of LLMs, it is challenging to adapt them to ASTE due to several reasons.
Notably, black-box models like GPT-4 \cite{gpt4-report} are less trustworthy and interpretable as it is not clear how to estimate the confidence of their predictions.
For instance, as each text may contain multiple sentiment triplets, it is useful to know which of the predicted triplets have higher confidence or lower confidence.
Hence, the lack of interpretability hinders the trustworthiness of LLMs in practical applications, and limits in-depth analysis of their performance.
On the other hand, it is generally not possible or feasible to train LLMs for specific tasks, leading to greater focus on prompt-based methods to improve performance.

Thus, we introduce confidence-aware sentiment extraction (\modelname{}), a simple and effective decoding strategy to improve the trustworthiness and performance of LLMs for complex sentiment tasks like ASTE.
Inspired by self-consistency \cite{wang2023selfconsistency} which samples diverse reasoning paths to select the most consistent answer, we sample diverse sets of sentiment triplets to select the most consistent triplets.
Intuitively, sentiment triplets which are most consistent, i.e., occur most often when sampling diverse sets of triplets, can be assigned a higher confidence.
Notably, it is simple to integrate \modelname{} with any language model that supports stochastic sampling, and it does not require any model re-training or access to model logits.
Compared to conventional decoding methods such as greedy search or beam search, \modelname{} enhances interpretability by estimating the confidence of each predicted triplet, and improves performance by explicitly considering a larger pool of sentiment triplets. 

In summary, our main contributions include: 
(1) To evaluate ASTE methods more holistically, we propose a domain-expanded benchmark which covers in-domain and out-of-domain performance across diverse domains.
(2) We annotate more than 4000 samples for two new domains based on hotel and cosmetics product reviews to support the new benchmark.
(3) We propose \modelname{}, a simple and effective decoding strategy to enhance the trustworthiness and performance of LLMs for ASTE. 
Our experiments demonstrate its effectiveness across different models, tasks, and settings.

\section{Related Work}


\paragraph{Aspect-Based Sentiment Analysis}
While sentiment analysis is often considered at the sentence or document level, this approach cannot capture the fine-grained sentiment einformation and nuaces of human opinions \cite{Poria2020BeneathTT}. 
To this end, aspect-based sentiment analysis (ABSA) consists of many task which aim to reveal richer sentiment information by considering the specific opinions and aspect targets in natural language \cite{pontiki-etal-2014-semeval}. 
Early works on ABSA focused on extracting individual sentiment elements, such as aspect term extraction \cite{liu-etal-2015-fine}, opinion term extraction \cite{yang-cardie-2012-extracting}, or aspect sentiment classification \cite{dong-etal-2014-adaptive}.
On the other hand, compound ABSA tasks have been introduced to jointly address multiple subtasks, including ASTE \cite{peng2020knowing} and ASQP \cite{zhang-etal-2021-aspect-sentiment}. 
In this work, we focus on ASTE which has many established methods, yet has not been studied through the lens of domain generalization \cite{Wang2021GeneralizingTU}.

\paragraph{Domain Generalization}

While traditional machine learning methods are trained based on the assumption that training and testing data are identically and independently distributed, this assumption seldom holds true in reality.
Hence, the performance of methods often deteriorates due to shifts in domain distributions \cite{Wang2021GeneralizingTU}.
As it is not feasible to comprehensively annotate task-specific data for training, there is an urgent need to improve the robustness and generalization ability of existing methods.
While are there many related topics such as domain adaptation \cite{7078994, gong-etal-2020-unified}, meta-learning \cite{Vilalta2002APV}, and lifelong learning \cite{Biesialska2020ContinualLL}, we believe that domain generalization is more widely applicable to the established methods for ASTE.
Hence, in this work, we mainly investigate domain generalization, the goal of which is to learn a model that will generalize well to unseen domains.

\paragraph{Large Language Models}

Recently, there have been numerous advancements in natural language processing due to the rapid development of large language models (LLMs) such as GPT-4 \cite{gpt4-report} and LLaMA \cite{Touvron2023LLaMAOA}.
Compared to the smaller pretrained language models (PLMs), LLMs have deeper language understanding and reasoning capabilities, owing to the large scale of the models and training data.
Moreover, the performance of LLMs can be further enhanced through methods such as instruction-tuning \cite{wei2022finetuned}, chain-of-thought prompting \cite{Wei2022ChainOT}, and reinforcement learning from human feedback \cite{Ouyang2022TrainingLM}.
However, there is less focus on fundamental decoding strategies that can heavily affect the behavior of generative methods.
On the other hand, language models are prone to hallucinating outputs that seem plausible but are incorrect or unreasonable \cite{Ji2022SurveyOH}, raising major concerns about their trustworthiness and interpretability \cite{Zhao2023ASO}.
Hence, we introduce a novel decoding strategy that aims to improve the performance and interpretability of LLMs for ASTE.

\section{Domain-Expanded ASTE Benchmark}

To evaluate the performance of ASTE methods more holistically and encourage development of more robust methods, we propose a domain-expanded benchmark.
The benchmark assesses models not only in-domain, but also in terms of out-of-domain generalization across diverse domains.
Hence, we construct the benchmark by leveraging two domains from existing datasets, while annotating samples for two new domains.
In this section, we detail the dataset construction process and dataset statistics for each domain. 

\subsection{Task Formulation}
\label{sec:formulation}

Given an input sentence $x$ containing $n$ words, ASTE aims to predict a set of sentiment triplets where each triplet $(t,o,p)$ corresponds to the aspect target, opinion, and sentiment polarity, respectively. 
Each aspect target $t$ and opinion $o$ are text spans in the sentence.
The sentiment polarity belongs to the label set of $\{\text{POS}, \text{NEG}, \text{NEU}\}$, which corresponds to positive, negative, and neutral sentiment, respectively. 

\subsection{Data Collection}
We construct a dataset with four domains by leveraging two domains from existing datasets \cite{peng2020knowing} and collecting data for two new domains.
Specifically, we collect review texts in the Hotel and Cosmetics domains from TripAdvisor Reviews \cite{angelidis-etal-2021-extractive} and Amazon Reviews \cite{he2016ups,mcauley2015image} respectively.
We collect 8000 samples from each domain corpus and use the spaCy tool to tokenize the review texts and label their part-of-speech tags.
To denoise the raw samples, we remove reviews that do not contain any nouns or adjectives.
We also leverage the existing Laptop and Restaurant domains from ASTE-Data-V2 \cite{xu-etal-2020-position}.
Within the Laptop and Restaurant domains, we remove duplicate samples and retain the existing triplet annotations.

\begin{table}[t]
    \centering
    \resizebox{1\columnwidth}{!}{
    \begin{tabular}{lcccc}
    \toprule
    Domain & Aspect Target & Opinion & Sentiment Triplet \\
    \midrule
    Hotel & 0.73 & 0.76 & 0.61 \\
    Cosmetics & 0.72 & 0.73 & 0.57 \\
    \bottomrule
    \end{tabular}
    }
    \caption{Inter-annotator agreement scores. We measure the agreement using the AvgAgr metric separately for aspect targets, opinions, and sentiment triplets.}
    \label{tab:agreement}
\end{table}

\setlength{\tabcolsep}{4pt} 
\begin{table}[ht]
    \centering
    \resizebox{1\columnwidth}{!}{
    \begin{tabular}{lcccccc}
    \toprule
        Domain & \#Train & \#Dev & \#Test & \#Triplets & \#T & \#O \\
    \midrule
    Restaurant & 1771 & 442 & 739 & 5376 & 1878 & 1743 \\
    Laptop & 867 & 217 & 362 & 2334&1086 &1083 \\
    Hotel & 1281 & 320 & 535 & 4064 & 1486 & 1706 \\
    Cosmetics & 1287 & 442 & 739 & 4002 & 1539 & 2221 \\
    
    \bottomrule
    \end{tabular}
    }
    \caption{
    Statistics of our domain-expanded ASTE dataset.
    We report the number of train samples, development samples, test samples, sentiment triplets, unique aspect targets (T), and unique opinions (O).
    }
    \label{tab:dataset_stat}
\end{table}

\subsection{Data Annotation}
For annotation, we follow the same data format as existing datasets \cite{peng2020knowing, xu-etal-2020-position}.
Specifically, annotators are provided with each tokenized review sentence as input.
They are required to annotate all valid sentiment triplets in the text according to the task formulation in Section \ref{sec:formulation}.
We include the detailed annotation guideline in the appendix.
To ensure the quality of data annotation, we conduct quality checking for each batch of annotated data.
Specifically, for each annotation batch, 10\% of the samples are randomly selected for manual checking. 
If more than 10\% of the selected samples contain errors, we provide detailed feedback and request annotators to amend the batch.
We engage two independent annotators to label the data and engage a third annotator to resolve any annotation disagreements.


Following previous works in data annotation for ABSA \cite{barnes-etal-2018-multibooked}, we measure the inter-annotator agreement using the AvgAgr metric \cite{Wiebe2005AnnotatingEO}:
\begin{align}
    \text{AvgAgr}(a, b) = \frac{1}{2} \left(
    \frac{| a \cap b |}{| a |} + 
    \frac{| a \cap b |}{| b |}
    \right)
\end{align}
\noindent where $a$ and $b$ are the set of annotations by the first and second annotators, respectively. 
Intuitively, the agreement value is the average of precision and recall between the two annotators.
Hence, the perfect agreement is 1 while no agreement is 0.
We report the inter-annotator agreement for the Hotel and Cosmetics domain in Table \ref{tab:agreement}.
We observe that the agreement scores are high and comparable to previous ABSA datasets \cite{barnes-etal-2018-multibooked}.

We report the statistics\footnote{We include more detailed analysis in Appendix \ref{sec:data_details}.}
of the domain-expanded dataset such as the number of reviews, sentiment triplets, and unique aspect targets in Table \ref{tab:dataset_stat}.

\begin{figure*}[!t]
\centering
\includegraphics[width=1.0\linewidth]{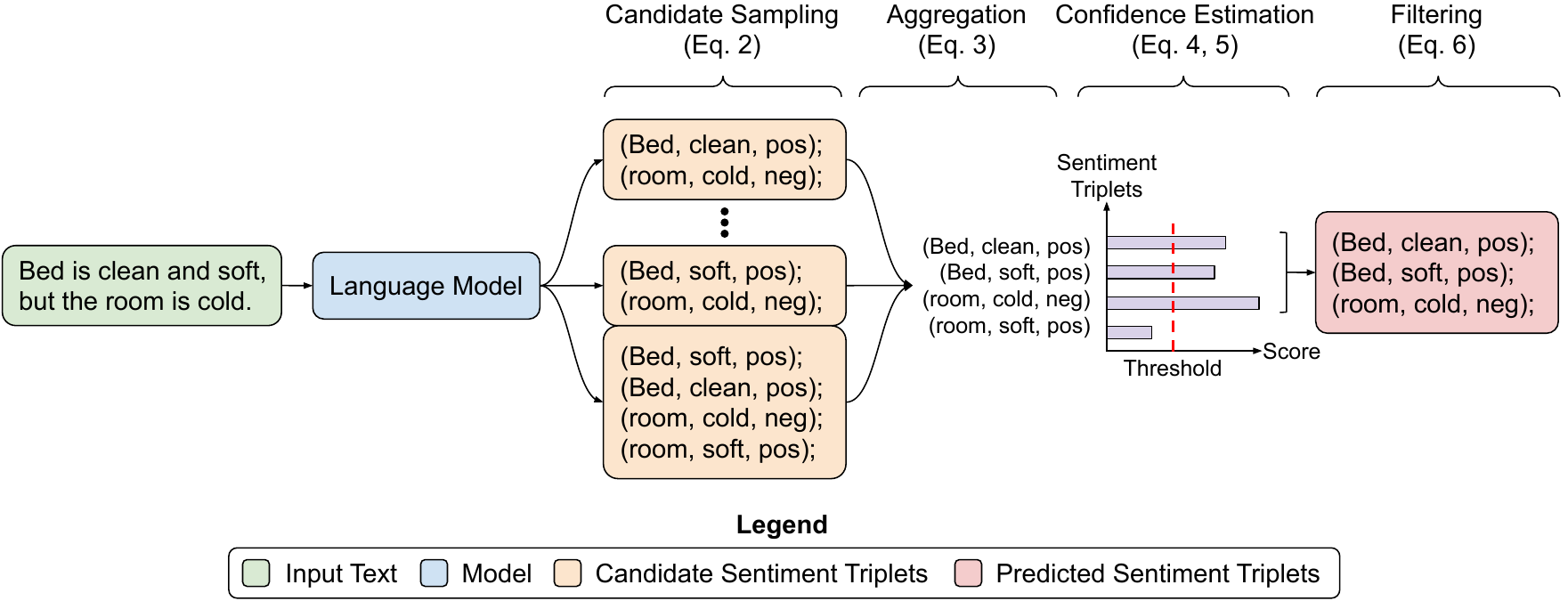}
\caption{
\flag{
Our proposed confidence-aware sentiment extraction (\modelname{}) decoding strategy which aims to enhance the trustworthiness and performance of LLMs for ASTE.
}
}
\label{fig:framework}
\end{figure*}

\section{Confidence-Aware Sentiment Extraction (\modelname{})} 
\flag{
To enhance the trustworthiness and effectiveness of large language models (LLMs) on ASTE, we propose confidence-aware sentiment extraction (\modelname{}), a simple and effective decoding strategy.
Compared to conventional decoding methods such as greedy search or beam search, \modelname{} enhances interpretability by estimating the confidence of each predicted triplet, and improves performance by explicitly considering a larger pool of sentiment triplets. 
Inspired by self-consistency \cite{wang2023selfconsistency} which samples diverse reasoning paths to select the most consistent answer, we sample diverse sets of sentiment triplets to select the most consistent triplets.
As shown in Figure \ref{fig:framework}, \modelname{} consists of four main steps: 
(1) Given the input text, we sample diverse output sequences from the language model, where each output sequence represents a set of candidate sentiment triplets.
(2) The unique sentiment triplets are then aggregated based on the sampled sets of triplets. 
(3) To estimate the confidence of each sentiment triplet, we calculate the occurrence frequency of each triplet.
(4) Lastly, we select the most confident sentiment triplets as the final predictions.
}

\subsection{Candidate Sampling}

In practice, generative methods such as sequence-to-sequence PLMs \cite{zhang-etal-2021-aspect-sentiment, zhang-etal-2021-towards-generative} and LLMs \cite{Wang2023IsCA, zhang2023reality} use approximate decoding methods such as greedy search or beam search as it is intractable to determine the optimal $y$ for a given input $x$, i.e., $\text{argmax}_y p(y \mid x)$.
Hence, we argue that generating a single sequence $y$ is sub-optimal as it only provides a narrow view of the possible triplet candidates.
On the other hand, sampling diverse sequences from the language model can provide the opportunity to consider a larger set of triplet candidates and estimate the confidence score of each triplet.
To obtain diverse triplet candidates, we use temperature-based sampling \cite{ficler-goldberg-2017-controlling, fan-etal-2018-hierarchical} which is a common method to generate diverse outputs from a language model.
Concretely, we sample $m$ outputs from our model $G$ for a given input $x$: 
\begin{align}
    S_j \sim G(x, k), \; j \in \{1, ..., m\}
\end{align}
\noindent
where $S_j$ denotes the set of sentiment triplets in the $j$-th sampled sequence.

\subsection{Aggregation}

Naturally, a triplet set may be sampled more than once and a sentiment triplet $(t, o, p)$ may be present in more than one set.
To aggregate the sentiment triplets, we take the union of the sampled sets to form the candidate set $S_c$:
\begin{align}
    S_c = \bigcup_{j=1}^{m} S_j
\end{align}
Hence, we only consider the unique sentiment triplets across all the sampled triplet sets.

\subsection{Confidence Estimation}

Intuitively, we assume that sentiment triplets that appear more frequently can be attributed to a higher confidence score.
Thus, we estimate the confidence score of each sentiment triplet $(t,o,p) \in S_c$ to be the corresponding occurrence frequency:
\begin{align}
    \phi(t, o, p) = \frac{\sum_{j=1}^{m}{\textbf{1}_{S_j}(t, o, p)}}{m}
\end{align}
where $\textbf{1}_{S_j}(t, o, p)$ is the indicator function of whether a triplet $(t, o, p)$ appears in $S_j$:
\begin{align}
    \textbf{1}_{S_j}(t, o, p) =
    \begin{cases}
    1 & \text{if $(t, o, p) \in S_j$} \\
    0 & \text{otherwise}
    \end{cases}
\end{align}

\noindent Naturally, the confidence score for each triplet is bounded within the range $0 \leq \phi(t, o, p) \leq 1$.
As it is not feasible to exhaustively sample from the language model, we sample $m=20$ output sequences for each input $x$. 

\subsection{Filtering}

While the steps thus far have improved interpretability through confidence estimation and triplet recall by sampling a larger pool of candidate triplets, we face the challenge noisy predictions.
Specifically, sampling more triplets may impact model precision due to increased numbers of false positive triplets.
Hence, we apply a confidence threshold $T$ over each triplet $(t,o,p) \in S_c$ to select the final prediction set $S_{\text{final}}$:
\begin{align}
    S_{\text{final}} = \{(t,o,p) \mid \phi(t,o,p) \geq T \}
\end{align}
This filtering process ensures that we retain only the higher-confidence triplets, thus mitigating noisy predictions.

\section{Experiment Setup}

\subsection{Settings}

In this work, we aim to provide a more holistic study of model performance on the ASTE task.
While previous works mostly focus on the in-domain setting, where the model is trained and tested on the same domain, we believe that this provides a limited perspective of model performance, as it does not consider robustness to domain shift.
Hence, we further evaluate models out-of-domain settings, where the model trained on one domain and tested on a different domain.
Moreover, certain models may be stronger in low-resource scenarios, which is important to consider as labeled data is often limited and costly to obtain in practice.
Thus, we further assess each model on the fully-supervised and few-shot scenarios.
Specifically, for the few-shot scenario, we sample 5 examples for each sentiment polarity.
Following previous works in ASTE \cite{peng2020knowing, xu-etal-2020-position}, we use the \metricname{} metric to measure model performance.
For all training experiments, we report the average results from 5 random runs.

\subsection{Models}

To provide a study of diverse models, we evaluate several ASTE methods based on pretrained language models (PLMs) and large language models (LLMs).
For PLMs, we include discriminative methods including GTS \cite{wu-etal-2020-grid} based on sequence tagging, Span-ASTE based on span enumeration and RoBMRC \cite{liu-etal-2022-robustly} based on machine reading comprehension.
We also consider generative methods including GAS \cite{zhang-etal-2021-towards-generative} and Paraphrase \cite{zhang-etal-2021-aspect-sentiment}.
As LLMs have shown general-purpose capabilities and strong performance on many language understanding and reasoning tasks, we also assess their performance on ASTE.
Specifically, we use the ChatGPT model API based on gpt-3.5-turbo-0301\footnote{\href{https://platform.openai.com/docs/models/gpt-3-5}{https://platform.openai.com/docs/models/gpt-3-5}}.
We note that while LLMs are technically PLMs as they also undergo large-scale pretraining, we use PLMs to refer to smaller models that are pretrained, such as BERT \cite{devlin-etal-2019-bert} and T5 \cite{JMLR:v21:20-074}. 
To adapt ChatGPT to complex sentiment tasks such as ASTE, we use in-context learning demonstrations \cite{wei2022emergent} with the prompt templates as shown in Appendix \ref{sec:prompt_templates}.
For the fully supervised scenario, we leverage in-context demonstration selection \cite{liu-etal-2022-makes} which selects relevant examples from the full dataset based on cosine similarity.
Specifically, we use embedding representations from Sentence-BERT \cite{reimers-gurevych-2019-sentence} and select the top-15 most similar examples as in-context demonstrations.
For the few-shot scenario, we use the few-shot examples as in-context demonstrations.

\subsection{Hyperparameters}

For all PLM-based methods, we use the base model size and original hyperparameters for training experiments.
For sampling with \modelname{}, we 
generate a fixed number of 10 outputs for each example.
To select the confidence threshold hyperparameter $T$, we perform a grid search with the values $\{0.0, 0.2, 0.4, 0.6, 0.8, 1.0\}$ based on \metricname{} results on the development set. 
For out-of-domain settings, we choose the confidence threshold from the respective source domain.
In addition, we report other experimental details in Appendix \ref{sec:hyperparameters}.

\section{Results and Analysis}

\flag{
To provide a holistic study of ASTE methods, we evaluate on the proposed domain-expanded ASTE benchmark, reporting the fully supervised in-domain results in Table \ref{tab:in_domain}, with fully supervised out-of-domain results in Table \ref{tab:out_domain}. 
We further study the few-shot scenario for in-domain and out-of-domain settings in Table \ref{tab:few_shot}.
In general, while specialized PLM-based methods currently outperform LLMs in the fully supervised scenario, there is a smaller performance gap for unseen domains, and LLMs exhibit better robustness to domain shift.
In contrast, we find that LLMs are more effective in low-resource scenarios, as evidenced by the few-shot results.
On the other hand, we observe that the proposed \modelname{} is an effective decoding strategy that not only addresses the fundamental interpretability limitation of LLMs, but also consistently improves performance across models, settings, and tasks.
}

\subsection{Fully Supervised Results} 
\label{sec:existing}

\begin{table*}[!t]
    \centering
    \resizebox{1\textwidth}{!}{
    \begin{tabular}{l ccc ccc ccc ccc|c}
    \toprule
    \multirow{2}{*}{{\textbf{Method}}}
    & \multicolumn{3}{c}{\textbf{Hotel}}
    & \multicolumn{3}{c}{\textbf{Laptop}}
    & \multicolumn{3}{c}{\textbf{Cosmetics}}
    & \multicolumn{3}{c}{\textbf{Restaurant}}
    & \multicolumn{1}{c}{\textbf{Avg.}} \\
    \cmidrule(lr){2-4}
    \cmidrule(lr){5-7}
    \cmidrule(lr){8-10}
    \cmidrule(lr){11-13}
    \cmidrule(lr){14-14}
    & $P.$ & $R.$ & \metricname
    & $P.$ & $R.$ & \metricname
    & $P.$ & $R.$ & \metricname
    & $P.$ & $R.$ & \metricname
    & \metricname \\
    \midrule
    GTS \cite{wu-etal-2020-grid} & 58.76 & 59.50 & 59.13 & 58.07 & 48.16 & 52.65 & 51.42 & 50.95 & 51.18 & 65.06 & 65.45 & 65.26 & 57.15 \\
    Span-ASTE \cite{xu-etal-2021-learning} & 67.73 & 62.92 & 65.24 & 60.73 & 54.40 & 57.39 & 59.79 & 55.0 & 57.29 & 68.69 & 65.41 & 67.01 & 61.74 \\
    RoBMRC \cite{liu-etal-2022-robustly} & 68.99 & 63.11 & 65.92 & 66.12 & 51.51 & 57.90 & 58.62 & 55.27 & 56.89 & 69.89 & 67.80 & 68.83 & 62.49 \\
    Paraphrase \cite{zhang-etal-2021-aspect-sentiment} & 65.21 & 61.07 & 63.08 & 61.23 & 55.13 & 58.02 & 58.45 & 53.62 & 55.93 & 68.56 & 68.46 & 68.51 & 61.41 \\
    GAS \cite{zhang-etal-2021-towards-generative} & 67.57 & 63.30 & 65.37 & 60.59 & 55.13 & 57.73 & 59.13 & 55.53 & 57.28 & 69.26 & 69.16 & 69.21 & 62.41 \\
    \hspace{3mm} with \modelname{} (Ours) & 67.40 & 64.75 & 66.05 & 60.60 & 56.79 & 58.63 & 59.51 & 57.01 & 58.23 & 68.84 & 70.42 & 69.62 & \textbf{63.13} \\
    \midrule
    ChatGPT & 47.59 & 53.13 & 50.20 & 44.57 & 49.12 & 46.74 & 34.80 & 38.73 & 36.66 & 53.49 & 57.68 & 55.50 & 47.28 \\
    \hspace{3mm} with \modelname{} (Ours) & 54.24 & 49.86 & 51.96 & 51.71 & 48.17 & 49.88 & 42.32 & 35.39 & 38.55 & 58.11 & 56.04 & 57.06 & \textbf{49.36} \\
    \bottomrule
    \end{tabular}
    }
    \vspace{-2mm}
    \caption{Evaluation results for \textbf{in-domain} ASTE with the full datasets.
    We report the average precision ($P$), recall ($R$), and \metricname scores for each domain, as well as the average \metricname{} (Avg.) across all domains. 
    }
    \label{tab:in_domain}
\end{table*}

\begin{table*}[!t]
    \centering
    \resizebox{1\textwidth}{!}{
    \begin{tabular}{l ccc ccc ccc ccc|c}
    \toprule
    \multirow{2}{*}{{\textbf{Method}}}
    & \multicolumn{3}{c}{\textbf{Hotel}}
    & \multicolumn{3}{c}{\textbf{Laptop}}
    & \multicolumn{3}{c}{\textbf{Cosmetics}}
    & \multicolumn{3}{c}{\textbf{Restaurant}}
    & \multicolumn{1}{c}{\textbf{Avg.}} \\
    \cmidrule(lr){2-4}
    \cmidrule(lr){5-7}
    \cmidrule(lr){8-10}
    \cmidrule(lr){11-13}
    \cmidrule(lr){14-14}
    & L$\rightarrow$H 
    & C$\rightarrow$H 
    & R$\rightarrow$H 
    & H$\rightarrow$L 
    & C$\rightarrow$L 
    & R$\rightarrow$L 
    & H$\rightarrow$C 
    & L$\rightarrow$C 
    & R$\rightarrow$C 
    & H$\rightarrow$R 
    & L$\rightarrow$R 
    & C$\rightarrow$R 
    & \metricname \\
    \midrule
    GTS \cite{wu-etal-2020-grid} & 35.05 & 52.75 & 49.41 & 34.01 & 32.68 & 40.98 & 38.08 & 24.31 & 32.77 & 55.73 & 49.86 & 49.94 & 41.65 \\
    Span-ASTE \cite{xu-etal-2021-learning} & 41.62 & 55.55 & 51.23 & 37.34 & 33.48 & 42.52 & 43.55 & 31.00 & 34.30 & 57.31 & 54.36 & 51.44 & 44.58 \\
    RoBMRC \cite{liu-etal-2022-robustly} & 36.17 & 58.17 & 52.67 & 37.77 & 35.57 & 41.26 & 41.81 & 26.97 & 32.12 & 60.47 & 51.10 & 55.73 & 44.76 \\
    Paraphrase \cite{zhang-etal-2021-aspect-sentiment} & 43.99 & 56.49 & 50.81 & 41.71 & 39.09 & 48.02 & 43.85 & 28.45 & 34.68 & 59.74 & 59.15 & 56.14 & 46.90 \\
    GAS \cite{zhang-etal-2021-towards-generative} & 46.18 & 59.10 & 52.71 & 40.77 & 37.88 & 48.25 & 46.10 & 29.81 & 34.97 & 59.57 & 60.47 & 56.54 & 47.76 \\
    \hspace{3mm} with \modelname{} (Ours) & 46.84 & 60.06 & 53.32 & 42.36 & 38.82 & 48.72 & 47.77 & 30.96 & 36.12 & 60.03 & 61.06 & 57.08 & \textbf{48.60} \\
    \midrule
    ChatGPT & 42.98 & 42.61 & 43.14 & 34.48 & 35.23 & 36.43 & 31.22 & 31.26 & 31.84 & 50.08 & 51.29 & 48.02 & 39.88 \\
    \hspace{3mm} with \modelname{} (Ours) & 42.91 & 45.07 & 45.56 & 36.08 & 36.66 & 38.57 & 31.04 & 31.60 & 32.80 & 51.78 & 53.74 & 50.43 & \textbf{41.35} \\
    \bottomrule
    \end{tabular}
    }
    \vspace{-2mm}
    \caption{Evaluation results for \textbf{out-of-domain} ASTE with the full datasets.
    We report the average \metricname score for each domain-pair (source domain $\rightarrow$ target domain), as well as the average \metricname{} (Avg.) across all domain-pairs.
    }
    \label{tab:out_domain}
\end{table*}

\begin{table}[ht]
    \centering
    \resizebox{1\columnwidth}{!}{
    \begin{tabular}{lccc}
    \toprule
    \textbf{Method} & \textbf{In-Domain \metricname{}} & \textbf{Out-Of-Domain \metricname{}} \\
    \midrule
    Span-ASTE & 32.65 & 20.71 \\
    Paraphrase & 33.46 & 22.95 \\
    GAS & 36.53 & 26.72 \\
    \hspace{3mm} with \modelname{} (Ours) & \textbf{38.42} & \textbf{28.81} \\
    \midrule
    ChatGPT & 44.38 & 38.19 \\
    \hspace{3mm} with \modelname{} (Ours) & \textbf{47.34} & \textbf{39.56} \\
    \bottomrule
    \end{tabular}
    }
    \caption{Evaluation results for \textbf{few-shot} ASTE (5-Shot). We report the average in-domain \metricname{} score across all domains, and the average out-of-domain \metricname{} score across all domain-pairs.
    }
    \label{tab:few_shot}
\end{table}

\paragraph{Evaluation of PLM-Based Methods}

\flag{
Based on the established methods that leverage PLMs, we find significant differences in performance and generalization for generative methods (i.e., Paraphrase, GAS) compared to discriminative methods (i.e., GTS, Span-ASTE, RoBMRC).  
Specifically, generative methods enjoy competitive in-domain performance and much stronger generalization to unseen domains, with an advantage  of more than 2 points in the out-of-domain setting on average.
Furthermore, while PLM-based methods generally demonstrate large performance disparities between in-domain and out-of-domain settings, generative methods are more robust to domain shift, as they exhibit smaller performance gaps on average (14.58) compared to discriminative methods (16.80).
We believe that this is largely due to the effect of label semantics \cite{ma-etal-2022-label}.
For instance, understanding that ``fresh'' is an adjective for describing food such as ``sushi'' in Figure \ref{fig:cover}, it can be easier for the model to predict the sentiment triplet (sushi, fresh, positive).
Hence, generative methods demonstrate better performance and generalization on the domain-expanded benchmark.
}

\paragraph{Comparison of LLM-Based Methods}

\flag{
By comparing the LLM-based ChatGPT to PLM-based methods, we observe that LLMs perform worse in general for fully-supervised scenarios, but show greater robustness to domain shift.
Notably, ChatGPT performs significantly worse on in-domain settings compared to PLM-based methods for ASTE. 
This is in contrast to their strong performance on simpler sentiment tasks such as sentence-level sentiment classification \cite{zhang2023reality}.
We believe that the difficulty that LLMs face in ASTE stems from the complexity of the task, as the structured nature of the sentiment triplets are less natural for language models.
Hence, there is larger area of improvement for task-specific adaptation of LLMs, especially for complex tasks such as ASTE.
On the other hand, we observe that ChatGPT can attain similar out-of-domain performance compared to some PLM-based methods, with a smaller performance gap between in-domain and out-of-domain settings (7.4).
We posit that the greater robustness to domain shift is due to exposure to more diverse pretraining data, which together with model scaling, imbues LLMs with comprehensive world knowledge \cite{safavi-koutra-2021-relational}.
This is consistent with previous findings that training data diversity is the main factor in robustness to domain shift \cite{NEURIPS2020_d8330f85}.
Thus, LLM-based methods show promising generalization to new domains, with ample room for future development.
}

\subsection{Few-Shot Performance}

In contrast to the fully supervised results, we find that LLMs show stronger performance in low-resource scenarios, as shown in Table \ref{tab:few_shot}.
Notably, ChatGPT significantly outperforms the PLM-based methods in both the in-domain and out-of-domain settings.
As LLMs benefit from massive scale of model parameters and training data, this enables them to learn a wider range of language patterns and semantics, hence generalizing well to new tasks, even with limited data \cite{NEURIPS2020_1457c0d6}.
From a practical point of view, while there remains ample room for improvement in the fully supervised scenarios, the strong generalization in low-resource scenarios and robustness to domain shift make LLMs suitable for data-scarce applications.
Hence, we believe that the few-shot results highlight the importance of evaluating ASTE methods on diverse scenarios, in order to provide a holistic view of their capabilities.

\begin{table*}[t!]
    \centering
    \resizebox{0.5\textwidth}{!}{
    \begin{tabular}{l|c|c|cc}
    \toprule
    \textbf{Task} & \textbf{Dataset} & \textbf{Method} & \textbf{Orig.} & \textbf{w/ \modelname{}} \\
    \midrule
    \multirow{4}{*}{AOPE}
    & Hotel & \multirow{4}{*}{GAS \cite{zhang-etal-2021-towards-generative}} & 71.77 & \textbf{72.44} \\
    & Laptop & & 65.93 & \textbf{66.77} \\
    & Cosmetics & & 62.98 & \textbf{63.91} \\
    & Restaurant & & 75.33 & \textbf{75.51} \\
    \midrule
    \multirow{2}{*}{ASQP}
    & Rest15 & \multirow{2}{*}{Paraphrase \cite{zhang-etal-2021-aspect-sentiment}} & 46.93 & \textbf{47.96} \\
    & Rest16 & & 57.93 & \textbf{58.86} \\
    \bottomrule
    \end{tabular}
    }
    \vspace{-2mm}
    \caption{Evaluation results for \textbf{in-domain} ABSA subtasks when using generative methods without change or with confidence-aware generative extraction (\modelname{}).}
    \vspace{-4mm}
    \label{tab:subtasks}
\end{table*}

\subsection{Impact of \modelname{}}

\flag{
While our proposed \modelname{} decoding strategy was mainly motivated by the limitations of interpretability and trustworthiness of black-box LLMs for ASTE, we find that it also provides reliable performance benefits.
Notably, we observe that ChatGPT with \modelname{} consistently outperforms the baseline which uses greedy decoding\footnote{While we have also experimented with beam search, we observed similar performance and hence used greedy search.} for both in-domain as well as out-of-domain settings.
Furthermore, as our decoding strategy is applicable to any method that supports stochastic sampling, we easily apply it to the generative method GAS \cite{zhang-etal-2021-towards-generative}, which also shows consistent benefits.
We believe that the performance benefits of \modelname{} stem mainly from the sampling process which considers more diverse sentiment triplets, which is supported by the significantly improved recall scores in Table \ref{tab:in_domain}.
On the other hand, there is little to no negative impact on precision, which suggests that our aggregation and filtering steps can effectively mitigate false positive triplets.
This is in contrast to conventional decoding methods such as greedy decoding, which only presents a single, less optimal set of sentiment triplets for consideration.
Hence, we believe that \modelname{} is an effective decoding strategy for ASTE and a promising direction for future development.
}

\subsection{Benefit of \modelname{} on Other ABSA Tasks}

As \modelname{} is a decoding strategy that can enhance the performance of generative models, it may also benefit other ABSA tasks.
Hence, to further study its effectiveness, we report the in-domain results of \modelname{}-based generative models for aspect opinion pair extraction (AOPE) \cite{chen-etal-2020-synchronous} and aspect sentiment quad prediction (ASQP) \cite{zhang-etal-2021-aspect-sentiment}.
We use our domain-expanded dataset for AOPE and the original Rest15 and Rest16 datasets for ASQP.
To modify our method for AOPE and ASQP, we simply consider pair sets and quadruplet sets respectively in the sampling process instead of triplet sets for ASTE.
Note that our method does not affect model parameters or re-training any models to be re-trained.
Based on the results in Table \ref{tab:subtasks}, we observe consistent improvement when using generative methods with \modelname{} compared to using the original greedy
decoding.
Furthermore, it can improve the interpretability and trustworthiness of generative ABSA predictions by estimating the confidence score of each pair, triplet, or quadruplet.
Overall, we believe that \modelname{} can be a beneficial and widely applicable technique for different ABSA tasks.





\subsection{Effect of Confidence-Aware Threshold}

As \modelname{} aims to improve the model recall while reducing false positives, it is crucial to remove the low-confidence triplets by applying a sufficiently high threshold filter.
However, a threshold that is too high may introduce more false negative triplets.
Hence, we study the effect of the confidence-aware threshold $T$ on model performance in Figure \ref{fig:threshold}.
We find that the in-domain performance is relatively stable across a wide range of thresholds between 0.2 and 0.8.
This suggests that the false positive triplets mainly have very low confidence scores i.e., $\phi(t,o,p) < 0.2$.
However, there is a sharp decrease in performance for extremely low or high threshold values, which is consistent with our intuition.

\begin{figure}
\centering
\resizebox{0.7\linewidth}{!}{
\begin{tikzpicture}
\pgfplotsset{width = 6cm, height = 4cm}
    \begin{axis}[
        ymax=70,
        ymin=45,
        ylabel={$F_1$ (\%)},
        label style={font=\fontsize{7}{1}\selectfont},
        xtick = {1,2,3,4,5,6},
        xticklabels = {0.0,0.2,0.4,0.6,0.8,1.0},
        xticklabel style = {font=\fontsize{7}{1}\selectfont},
        yticklabel style = {font=\fontsize{7}{1}\selectfont},
        xtick pos = left,
        ytick pos = left,
        ymajorgrids = true,
        grid style=dashed,
    ]
    \addplot [mark=square, mark size=1.2pt, color=orange] plot coordinates {
    (1, 51.9749) 
    (2, 62.9272) 
    (3, 64.547) 
    (4, 64.4688) 
    (5, 62.6802)
    (6, 54.9022)
    };
    \end{axis}
\end{tikzpicture}
}
\vspace{-3mm}
\caption{The effect of confidence-aware threshold $T$ on in-domain performance for the Hotel domain.}
\vspace{-2mm}
\label{fig:threshold}
\end{figure}

\section{Conclusions}
In conclusion, this work addressed the task of Aspect Sentiment Triplet Extraction (ASTE) in sentiment analysis, focusing on the issues of limited benchmark domains and the challenges of large language models (LLMs) in handling complex sentiment tasks. 
We introduced a domain-expanded ASTE benchmark by annotating samples from diverse domains, enabling the evaluation of models in both in-domain and out-of-domain settings. This expanded benchmark provided a more comprehensive assessment of model performance, addressing concerns regarding the generalizability of proposed methods.
Secondly, a novel decoding strategy called CASE (Context-Aware Sampling and Enhancement) was proposed to enhance the trustworthiness and performance of LLMs in ASTE. 
The experimental results demonstrated its effectiveness across multiple tasks, settings, and models. Its simplicity and efficacy make it a promising general decoding strategy for complex sentiment tasks.
By expanding the scope of evaluation and providing a reliable decoding strategy, we hope to encourage the research community to rethink the generalizability of benchmarks and models for ASTE. The findings highlight the importance of considering diverse domains and utilizing appropriate decoding strategies when tackling fine-grained sentiment analysis tasks. With these contributions, we hope to foster the development of more robust and capable sentiment analysis methods in the future.


\section*{Acknowledgment}

This work was substantially supported by DAMO Academy through DAMO Academy Research Intern Program.

\section*{Limitations}

As our method samples multiple output sequences for a given input sequence, there is an increased computational cost for inference. 
However, this is a trade-off similar to tuning hyperparameters for beam search in text generation problems, and the effect can be mitigated by batched inference.
Our method also relies on the sampled sequences to have sufficient diversity in other to consider a larger set of candidate triplets. 
However, too much diversity may introduce unwanted noise.
The diversity is affected by both the temperature sampling hyperparameter and the number of sampled sequences. 
In this work, we keep the temperature sampling hyperparameter fixed as a standard value for generation due to computational constraints.
We analyze the effect of the number of sampled sequences $m$ in Appendix \ref{sec:sampling}.

\section*{Ethics Statement}

For data annotation, we engage two professional annotators who are fairly compensated. 
The compensation is negotiated based on the task complexity and assessment of a reasonable annotation speed.
The annotators have given their consent for their annotations to be publicly released as a research dataset.
The data annotation project pass the ethics review of the data annotation team as it does not contain any confidential data.
The data annotators are adults who are versed in multiple languages.
We release our datasets under the same license (CC BY NC 4.0) as the original data that we collected from.
The licence allows for free sharing and adaptation of the dataset as long as appropriate credit is given, and the data is only used for non-commercial purposes.

\bibliography{anthology,custom}
\bibliographystyle{acl_natbib}

\appendix

\section{Appendix}

\subsection{Duplicate-Aware Evaluation for ASTE}
\label{sec:scoring}

\begin{algorithm}
\small
\SetAlgoLined

\PyCode{} \\
\PyCode{num\_pred = 0} \PyComment{Count of predicted triplets} \\
\PyCode{num\_gold = 0} \PyComment{Count of gold triplets} \\
\PyCode{num\_correct = 0} \PyComment{Count of correct triplets} \\

\PyCode{} \\
\PyComment{Match predicted and gold triplets} \\
\PyComment{using set intersection} \\
\PyCode{for sentence in data:}  \\
\Indp   
    \PyCode{pred\_set = set(sentence.pred\_triplets)} \\
    \PyCode{gold\_set = set(sentence.gold\_triplets)} \\
    \PyCode{correct\_set = pred\_set \& gold\_set} \\
    \PyCode{} \\
    \PyCode{num\_pred += len(pred\_set)} \\
    \PyCode{num\_gold += len(gold\_set)} \\
    \PyCode{num\_correct += len(correct\_set)} \\

\Indm 
\PyCode{} \\
\PyComment{Calculate scores} \\
\PyCode{p = num\_correct / num\_pred)} \PyComment{Precision} \\
\PyCode{r = num\_correct / num\_gold)} \PyComment{Recall} \\
\PyCode{f1\_score = 2 * p * r / (p + r)} \\

\Indm
\caption{
Pseudocode of duplicate-aware Micro-\metricname{} evaluation for ASTE.
}
\label{algo:scoring}
\end{algorithm}

\subsection{Additional Hyperparameters}
\label{sec:hyperparameters}

\begin{table}[!t]
    \centering
    \resizebox{0.9\columnwidth}{!}{
    \begin{tabular}{lr}
    \toprule
    Name & Value \\
    \midrule
    GPU Model & Nvidia A6000 \\
    CUDA Version & 11.3 \\
    Python Version & 3.7.12 \\
    PyTorch Version & 1.11.0 \\
    ChatGPT API Cost & \$110 \\
    Generation Sampling Temperature & 1.0 \\
    \bottomrule
    \end{tabular}
    }
   \caption{List of experimental details.} 
    \label{tab:hparams}
\end{table}

For GAS and Paraphrase models, there are 140M parameters when using BART-base. When using T5, there are 220M parameters. For BERT-base models (GTS, Span-ASTE, RoBMRC), there are roughly 110M parameters.

\subsection{Effect of Sampling Size}
\label{sec:sampling}

For sampling number of sequence $m$, there are on average 3.12, 3.45, 3.84 unique triplets sampled for $m=10,20,30$ respectively.

\subsection{Annotation Guide}
\label{sec:annotation_guide}
This section illustrates the guideline for human annotators. This task is a fine-grained sentiment analysis task where opinion terms, their aspect targets, and their expressed sentiments should be extracted together. Each sample contains one or multiple sentences which have been tokenized and labeled with indices. The annotation steps are as follows: 
\begin{enumerate}
    \item Read and understand the text sample and find out opinion terms as well as aspect target terms. Note that these terms should be explicit and the target term should not be a pronoun. If there is no opinion term or aspect target term, the sample is marked as ``Invalid''.
    \item If the sample contains opinion terms and aspect target terms, check whether there are aspect-opinion pairs. If not, the sample should also be marked as ``Invalid''.
    \item Determine the expressed sentiment of these pairs and record the spans of aspect-opinion pairs and their expressed sentiment in a 3-tuple format. Note that each sentence can have multiple triplets. 
\end{enumerate}

For example, given a review ``The room was huge but terribly furnished''. We can find two aspect-opinion pairs (room, huge) with positive sentiment and (room, terribly furnished) with negative sentiment. The triplets of this text sample should be recorded in this format: ([1], [3], ``POS''), ([1], [5, 6], ``'NEG'), where the index of the first token is 0. 

There are several special cases that may make annotators hard to determine. We give a uniform guide here:
\begin{itemize}
    \item Articles such as ``the'', ``a'', and ``an'' should not be included in target terms.
    \item Separate conjoined terms. For example, ``The bedroom and washroom are big and clean''. ``Bedroom and washroom'' should be recorded as two separate terms ``bedroom'' and ``washroom''. Opinion terms ``big'' and ``clean'' should also be separated.
    \item It might be hard to determine whether some adverbs should be included in opinion terms. We should include these adverbs if they have a large influence on the sentiment polarity of the opinion term. For example, ``This room is too big.'' The opinion term should be ``too big'' instead of ``big'', since ``too'' makes the opinion term express an obvious negative sentiment.
\end{itemize}

\subsection{Detailed Results}
\label{sec:detailed_results}

\subsection{Prompt Templates}
\label{sec:prompt_templates}
To adapt ChatGPT to complex sentiment tasks such as ASTE, we design several templates based on previous works in generative ASTE \cite{zhang-etal-2021-aspect-sentiment}.

\subsection{More Details of Datasets}
Table \ref{sec:data_details} shows more details of our domain-expanded ASTE dataset. We can observe that our annotated hotel and cosmetics domains contain a larger average sample length and their label distribution is more balanced than previous restaurant and laptop domains.

\label{sec:data_details}
\setlength{\tabcolsep}{4pt} 
\begin{table}[!t]
    \centering
    \resizebox{1\columnwidth}{!}{
    \begin{tabular}{l|rrrr}
    \toprule
        Domain & Average Sample Length & POS\% & NEU\% & NEG\% \\
    \midrule
    Restaurant & 16.37 tokens & 73.01\% & 6.75\% & 20.24\%  \\
    Laptop & 18.36 tokens & 57.50\% & 9.64\% & 32.86\% \\
    Hotel & 21.92 tokens & 59.25\% & 11.69\% & 29.06\%  \\
    Cosmetics & 21.61 tokens & 45.68\% & 25.59\% & 28.74\% \\
    
    \bottomrule
    \end{tabular}
    }
    \caption{
    More details of our domain-expanded ASTE dataset.
    We report the average length of samples and the percentage of positive (POS\%), neutral (NEU\%) and negative (NEG\%) triplets respectively.
    }
    \label{tab:dataset_details}
\end{table}

\subsection{Dataset Examples}

Table \ref{tab:dataset_examples} presents five examples for each domain. The standard of triplet formulation is the same across four domains and aspect target terms are domain-specific, indicating that our domain-expanded dataset can be well used as a cross-domain ASTE benchmark.

\begin{table*}[ht]
    \centering
    \resizebox{1\textwidth}{!}{
    \begin{tabular}{l|ll}
    \toprule
       \textbf{Domain} & \textbf{Example}  & \textbf{Triplets} \\
    \midrule
        \multirow{5}{*}{Restaurant} & The service is awful . & (service, awful, negative) \\
        & The chicken dinner was real good . &(chicken dinner, good, positive) \\
        &The food is reliable and the price is moderate . & (food, reliable, positive), (price, moderate, neutral) \\
        &Staffs are not that friendly , but the taste covers all . & (staffs, not that friendly, negative), (taste, covers all, positive) \\
        &Prices are in line . & (prices, in line, neutral) \\
    \midrule
        \multirow{5}{*}{Laptop} & The keyboard feels good and I type just fine on it . &  (keyboard, good, positive) \\
        & The battery gets so HOT it is scary . & (battery, HOT, negative), (battery, scary, negative) \\
        & It 's great for streaming video and other entertainment uses .& (streaming video, great, positive), (entertainment uses, great, positive)\\
        &This mouse is terrific . & (mouse, terrific, positive) \\
        & Of course my warranty runs out next month . & (warranty, runs out, neutral) \\
    \midrule
       \multirow{5}{*}{Hotel} & The smell was only slightly less prominent in our corner suite at the end of the hallway . & (smell, prominent, neutral) \\
       & Also , the garbage trucks that frequent the ally are loud . & (garbage trucks, loud, negative) \\
       & In the morning you can enjoy a free breakfast with many choices . & (breakfast, enjoy, positive), (breakfast, free, positive) \\
       & The price was reasonable compared to the other options in the area . & (price, reasonable, positive) \\
       & My fiancé opened the window shades and we had a huge brick wall for a view . & (brick wall, huge, neutral)\\
    \midrule
        \multirow{5}{*}{Cosmetics} & It use to be one of the best products in the market . & (products, best, positive) \\
        &This is a very heavy cover - up that feels heavy on your face .  & (cover-up, heavy, neutral) \\
        & Flimsy is really not a great thing when it 's 20 bucks . & (Flimsy, not a great thing, negative) \\
        & I ordered the blonde color , but it really is a little dark . & (color, blonde, neutral), (color, dark, neutral) \\
        &I love Essie but the formula on this one is awful . & (Essie, love, positive), (formula, awful, negative) \\
    \bottomrule
    \end{tabular}
    }
    \caption{Dataset examples.}
    \label{tab:dataset_examples}
\end{table*}

\subsection{Case Study}
Table \ref{tab:case_study} compares predictions of GAS and our GAS+CAGE method on two examples in two cross-domain settings. We find both methods show great performance in determining the sentiment. However, our method can identify the number of triplets more correctly, indicating that CAGE can effectively mitigate pseudo-label noise by reducing false positives and false negatives.

\begin{table*}[ht]
    \centering
    \resizebox{1\textwidth}{!}{
    \begin{tabular}{l|ll}
    \toprule
        & \textbf{Hotel -> Cosmetics}  & \textbf{Cosmetics -> Hotel} \\
    \midrule
        \multirow{3}{*}{Example} & Though it is more expensive than mass market gels , & The rooms were very clean and the staff was very friendly \\
        & it does provide higher performance . & and helpful especially when it came to ensuring we got on \\
        & & our buses for tours and our flights back home . \\
    \midrule
        \multirow{2}{*}{Gold label} & \multirow{2}{*}{(performance, higher, positive)} &  (rooms, clean, positive), (staff, friendly, positive), \\
        & & (staff, helpful, positive) \\
    \midrule
       \multirow{2}{*}{GAS prediction} & (performance, higher, positive), & (rooms, clean, positive), \\
       & (gels, expensive, negative) & (staff, friendly, positive) \\
    \midrule
        \multirow{2}{*}{GAS+CAGE prediction} & \multirow{2}{*}{(performance, higher, positive)} & (rooms, clean, positive), (staff, friendly, positive), \\
        & & (staff, helpful, positive) \\
    \bottomrule
    \end{tabular}
    }
    \caption{Case Study.}
    \label{tab:case_study}
\end{table*}

\end{document}